# Convolutional Neural Network Compression Based on Low-Rank Decomposition

Yaping He
*College of Computer and Information Science College,
Southwest University,
Chongqing, China*
yapinghe93@gmail.com

Linhao Jiang
*Shapingba Community Health Service Center,
Shapingba District,
Chongqing, China*
18333140@qq.com

Di Wu
*College of Computer and Information Science College,
Southwest University,
Chongqing, China*
wudi.cigit@gmail.com

*Abstract*—Deep neural networks typically impose significant computational loads and memory consumption. Moreover, the large parameters pose constraints on deploying the model on edge devices such as embedded systems. Tensor decomposition offers a clear advantage in compressing large-scale weight tensors. Nevertheless, direct utilization of low-rank decomposition typically leads to significant accuracy loss. This paper proposes a model compression method that integrates Variational Bayesian Matrix Factorization (VBMF) with orthogonal regularization. Initially, the model undergoes over-parameterization and training, with orthogonal regularization applied to enhance its likelihood of achieving the accuracy of the original model. Secondly, VBMF is employed to estimate the rank of the weight tensor at each layer. Our framework is sufficiently general to apply to other convolutional neural networks and easily adaptable to incorporate other tensor decomposition methods. Experimental results show that for both high and low compression ratios, our compression model exhibits advanced performance.

## I. INTRODUCTION

Given the rapid development of artificial intelligence technology [1]-[10], deep neural networks have achieved great success. They have been widely used in tasks such as image classification [11], object detection [12], and so on. Nevertheless, executing DNNs in embedded devices is still challenging. Many of the resource-constrained devices face the challenge that advanced DNNs have parameter sizes too large for real-world deployment. This limitation hinders the practical deployment of the models. In response to this challenge, numerous researchers have proposed various types of strategies to lightweight models and reduce model parameters.

**Tensor Decomposition for Model Compression.** Tensor decomposition [13]-[20] offers significant advantages in managing large-scale tensor data [20]-[26] owing to its favorable mathematical properties. As such, tensor decomposition has emerged as an exceedingly appealing technique for model compression. It can offer exceptionally high compression ratios for models with minimal performance degradation compared to other compression schemes [1]. In particular, for recurrent neural networks [27][28], state-of-the-art tensor decomposition methods can achieve thousands of times parameter compression, while significantly enhancing accuracy. Driven by the impressive compression performance, an increasing number of researchers are investigating model compression [29][30] methods rooted in tensor decomposition.

**Limitations of prior art.** Despite the maturity of tensor decomposition techniques and the satisfactory results obtained, there are limitations. Parametric compression achieving thousands of times reduction is only prevalent in video compression tasks. In the case of Convolutional Neural Networks (CNNs) employed in image categorization tasks, performance severely deteriorates, even with state-of-the-art Tensor Train (TT) and Tensor Ring (TR) decompositions. For instance, despite leveraging the latest advances in TR, a 1.0% decrease in accuracy persists with a compression ratio of 2.7 for the ResNet-32 model on the CIFAR-10 dataset. With a larger compression ratio of 5.8, the accuracy loss escalates to 1.9% [24].

**What limits performance?** Tensor decomposition suffers from the above limitations mainly due to the model's lack of robustness and the choice of rank. Specifically, tensor decomposition typically breaks down a layer of a neural network into multiple consecutive sublayers, and the resulting factor matrix often adheres to specific mathematical forms, such as orthogonality [32][33]. However, the factor matrix, representing the weight matrix of the compressed neural network's sublayers, gets updated during each iteration, making it challenging to satisfy the initial mathematical properties. Additionally, a significant imbalance exists between the input and output channels in the neural network, rendering decomposing different modalities using the same rank highly unreasonable [34]. In short, these two aforementioned factors can significantly compromise the model's performance.

**Technical Preview and Contributions.** To address these limitations, our paper introduces a model compression method that integrates orthogonal regularization and VBMF. Initially, the model undergoes training with over-parameterization while imposing orthogonal regularization simultaneously, ensuring it achieves or exceeds the accuracy of the original model. Subsequently, VBMF is utilized to estimate the tensor rank for each layer of the neural network. Finally, low-rank training is conducted, leading to the acquisition of the compressed model. In summary, the contributions of this paper are outlined as follows:

- We propose a framework for addressing the model compression problem. By conducting training with over-parameterization and orthogonal regularization, not only are better initial values provided for low-rank models, but also orthogonal properties are obtained.
- We employ the VBMF method to estimate the rank of one modality in the TK-2 decomposition, while the other modality is addressed through the relationship among the input and output channels of the convolutional neural network.
- The proposed framework is evaluated across various DNN models employed in image classification. Experimental results show that for both high and low compression ratios, our compression model exhibits advanced performance.

II. RELATED WORK ON DNN MODEL COMPRESSION

*A. Sparsification*

Sparsification usually refers to reducing the number of model parameters in machine learning and computer vision to improve the efficiency and performance of the model. In deep learning, sparsification techniques[35]-[42] can help reduce the complexity of the model, lowering memory and computational requirements while potentially improving the generalization of the model. Different levels of network structures can be sparse, such as weights, filters, and channels. To obtain sparsity, models can be trained with pruning or sparse perceptual regularization. Sparsification can also be classified as structured [43] and unstructured [44]. Unstructured models usually have high accuracy and compression ratios, but this approach is not friendly to embedded platforms due to irregular memory accesses and unbalanced workloads [45][46]. Structured sparse models are more adapted to hardware deployments, however their compression performance is usually not as good as the same type of unstructured models.

*B. Tensor Decomposition*

Tensor decomposition is a widely used technique in machine learning and data science that simplifies models and reduces computational complexity by decomposing higher-order tensors into products of lower-order tensors. Truncated singular value decomposition (TSVD) has good performance in decomposing matrices, however, compared to other tensor decomposition methods such as Tucker [47], and CANDECOMP/PARAFAC (CP) decomposition [48], TSVD [49] is slightly weak when dealing with higher order tensors, which can result in severe loss of accuracy with limited compression ratio. With the development of tensor network technology, TT [50]-[55], and TR decomposition have become the mainstream of compression methods. Due to their unique structural properties, they can give the model an extremely high compression ratio. These advantages are more obvious in RNN networks for video tasks.TT and TR are used to decompose the network input and hidden layers, and it is even able to compress the network tens of thousands of times in some special networks. However, TT [50] and TR [56] perform poorly in some CNNs and cause severe accuracy loss even with a small compression ratio. From a practical application point of view, this non-negligible accuracy loss limits the wide use of the model in embedded devices.

*C. Rank Selection*

Tensor decomposition techniques can significantly reduce the number of parameters and computational complexity of a model by decomposing a high-dimensional tensor into a sum of low-rank tensors when compressing a neural network model [58]-[62]. Selecting the optimal rank is the key to achieving effective compression, which requires a balance between model accuracy and computational efficiency. Auto-ML methods [63], automated techniques based on parameter estimation and genetic algorithms [64]-[67], and variational Bayesian [68]-[70] matrix decomposition have been proposed to automate the selection of the optimal rank. In addition, the practice involves manually adjusting and experimenting with different rank choices, as well as approximating matrices using techniques such as TSVD. Together, these methods have contributed to the application and development of tensor decomposition techniques in deep-learning model compression.

III. MODEL COMPRESSION METHOD

*A. Low-rank Tucker representation of Convolutional layers*

In CNNs, Convolutional layers map an input tensor $\mathbf{X} \in \mathbb{R}^{S \times W \times H}$ into an output tensor $\mathbf{Y} \in \mathbb{R}^{T \times W' \times H'}$ with a four-order kernel tensor $\mathbf{K} \in \mathbb{R}^{D \times D \times S \times T}$. Convolutional computation represents a linear mapping. And it is defined as [47]:

$$\mathbf{Y}(w',h',t) = \sum_{i=1}^{D}\sum_{j=1}^{D}\sum_{s=1}^{S} \mathbf{K}_{i,j,s,t} \mathbf{X}_{h_i,w_i,s}, \qquad (1)$$
$$h_i = (h'-1)\Delta + i - P \text{ and } w_j = (w'-1)\Delta + j - P,$$

where the size of $\mathbf{K}$ is $D \times D \times S \times T$, $\Delta$ and $P$ represent stride and zero-padding size, respectively.

TK decomposition has unique advantages in CNNs, so we will use TK decomposition to decompose the convolutional layers in this paper. The decomposition process is as follows:

$$K_{i,j,s,t} = \sum_{r_1=1}^{R_1}\sum_{r_2=1}^{R_2}\sum_{r_3=1}^{R_3}\sum_{r_4=1}^{R_4} G_{r_1,r_2,r_3,r_4} U^{(1)}_{i,r_1} U^{(2)}_{j,r_2} U^{(3)}_{s,r_3} U^{(4)}_{t,r_4}, \qquad (2)$$

where $G \in \mathbb{R}^{R_1 \times R_2 \times R_3 \times R_4}$ is core tensor, $U^{(1)} \in \mathbb{R}^{D \times R_1}$, $U^{(2)} \in \mathbb{R}^{D \times R_2}$, $U^{(3)} \in \mathbb{R}^{S \times R_3}$ and $U^{(4)} \in \mathbb{R}^{T \times R_4}$ are factor matrices.

In TK decomposition, not every mode needs to be decomposed. For example, weight K can retain mode-1 and mode-2 because the size of the convolution kernel will usually be small, such as 3×3 or 5×5. Thus, the weight tensor can be expressed as follows:

$$K_{i,j,s,t} = \sum_{r_3=1}^{R_3}\sum_{r_4=1}^{R_4} G_{i,j,r_3,r_4} U^{(3)}_{s,r_3} U^{(4)}_{t,r_4}, \qquad (3)$$

where $G \in \mathbb{R}^{D \times D \times R_3 \times R_4}$ is a core tensor. This method is known as TK-2 decomposition. Considering the advantages of convolutional operations in terms of parallelism, etc., we reconstruct the factor matrix into three consecutive convolutional layers in this paper.

$$Z_{h,w,r_3} = \sum_{s=1}^{S} U^{(3)}_{s,r_3} X_{h,w,s}, \qquad (4)$$

$$Z'_{h',w',r_4} = \sum_{i=1}^{D}\sum_{j=1}^{D}\sum_{r_3=1}^{R_3} G_{i,j,r_3,r_4} Z_{h_i,w_j,r_3}, \qquad (5)$$

$$Y_{h',w',t} = \sum_{r_4=1}^{R_4} U^{(4)}_{t,r_4} Z'_{h',w',r_4}, \qquad (6)$$

where $Z$ and $Z'$ are the intermediate output tensors. The above process is equivalent to a three-layer convolutional network without linear transformations, where the first convolutional layer from $S$ feature maps to $R_3$ feature maps with 1×1 kernel size, the second convolutional layer from $R_3$ feature maps to $R_4$ feature maps with $D \times D$ kernel size, the last convolutional layer from $R_4$ feature maps to $T$ feature maps with 1×1 kernel size. This substructure is more common in other classical network structure, such as the inception module in GoogleNets, where there are no nonlinear layers between the 1×1 and $D \times D$ convolutional layers. The convolutional mapping process of TK-2 is shown in Figure 1.

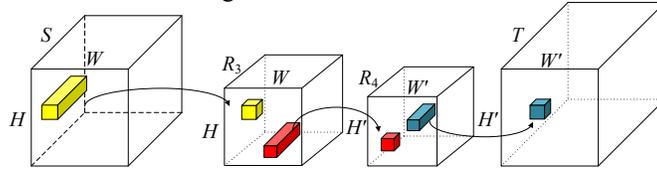

Fig. 1. TK-2 decompositions for compact a convolution.

### B. Low-rank TSVD representation of Fully Connected layers

For classical neural networks, the convolutional layer is a fourth-order tensor and the Fully Connected (FC) layer is a second-order matrix. The FC layer is computed as follows:

$$y^T = x^T W, \qquad (7)$$

where $x \in \mathbb{R}^M$ is an input vector, $y \in \mathbb{R}^N$ is an output vector, and $W \in \mathbb{R}^{M \times N}$ is a weight matrix. Since the weights are in matrix form and their order is not high, we use TSVD to compress the FC layer instead of CP or TK-2 decomposition. TSVD can decompose the weight matrix into three parts, which are the left singular matrix $U \in \mathbb{R}^{M \times R}$, the right singular matrix $V^T \in \mathbb{R}^{R \times N}$, and the diagonal singular value matrix $S \in \mathbb{R}^{R \times R}$. $R$ represents weight matrix rank. The larger the $R$ the larger the compression ratio, but the more the model will lose performance. The decomposition process is as follows:

$$W = USV^T, \qquad (8)$$

Integrate the computational processes of the FC layer and TSVD with each other. It is easy to obtain the following form:

$$\mathbf{y}^T = (\mathbf{x}^T (\mathbf{US}))\mathbf{V}^T, \tag{9}$$

This means that a layer of full connectivity can be replaced by two consecutive sublayers. It can be expressed as follows:

$$\mathbf{z}^T = \mathbf{x}^T (\mathbf{US}), \tag{10}$$

$$\mathbf{y}^T = \mathbf{z}^T \mathbf{V}^T. \tag{11}$$

*C. Theoretical complexity analysis*

The original convolution operation and FC layer require model parameters for $D^2ST$ and $MN$ and bring about computational consumption with $H'W'D^2ST$ and $MN$. After compression, the theoretical compression ratio (*CR*) and speed-up ratio (*SR*) of the convolutional layer can be obtained by

$$CR = \frac{TSD^2}{RS + RD^2 + TR}, \tag{12}$$

$$SR = \frac{TSD^2W'H'}{RSWH + RD^2W'H' + TRW'H'}, \tag{13}$$

For an FC layer, the computation process is similar to that of a convolutional layer. We can consider an FC layer as a $1\times 1$ convolutional layer.

$$CR = SR = \frac{MN}{MR + RN}. \tag{14}$$

---

**Algorithm 1**: Orthogonal regularization training
**Input:** Dataset $D$, ranks $\{R\}$, TK-2 decomposition for each layer
$\mathbf{G} \in \mathbb{R}^{R1\times R2\times R3\times R4}$ $\mathbf{U}^{(3)} \in \mathbb{R}^{S\times R3}$ and $\mathbf{U}^{(4)} \in \mathbb{R}^{T\times R4}$,
orthogonal parameters $\rho$, $\lambda$, training epochs $E$, $e$.
**Ouput:** $\mathbf{U}^{(3)}$, $\mathbf{U}^{(4)}$, $\mathbf{G}$.
**Initialize**: xavier_uniform($\mathbf{U}^{(3)}$, $\mathbf{U}^{(4)}$, $\mathbf{G}$).
// over-parameterized training
**for** $t=1$ to $E$ **do**
    $\mathbf{X}, \mathbf{Y} \leftarrow$ sample_batch($D$);
    $\hat{\mathbf{Y}} \leftarrow$ forward ($\mathbf{X}$, $\{\mathbf{U}^{(3)}, \mathbf{U}^{(4)}, \mathbf{G}\}$) via (4), (5) and (6);
    // Orthogonal regularization
    Calculate $L_{R3}$ via (15);
    Calculate $L_{R4}$ via (16);
    Calculate $L$ via (17);
    Updata ($\{\mathbf{U}^{(3)}, \mathbf{U}^{(4)}, \mathbf{G}\}$, loss);
**end**
// low-rank training
**for** $t$ to e **do**
    retraining ($\{\mathbf{U}^{(3)}, \mathbf{U}^{(4)}, \mathbf{G}\}$, loss);
**end**

---

*D. Orthogonal regularization for CNNs*

The factorized factor matrices satisfy orthogonality for both TK-2 and SVD, which provide sufficient theories guarantee for low-rank approximation. However, the factor matrix acts as a free variable in the decomposition process and it will be updated at each iteration. Therefore, its orthogonality is difficult to be guaranteed. It is essential to employ various optimization strategies [71]-[80].

To improve the performance of the compression model while reducing the computational and storage expenses, we adopt a high-rank regularization training strategy [81]-[85]. Specifically, the weight tensor is first decomposed with high rank, and the model is more honorably close to the accuracy of the original model due to its large number of parameters. Second, orthogonal regularization is applied to the factor matrix of the decomposition [86]-[90]. The rationale for training based on this idea is that orthogonality maximizes the ability to represent information. The orthogonal regularization is calculated as follows:

$$L_{R_3}(U^{(3)}) = \frac{\rho}{R_3}(\|U^{(3)T}U^{(3)} - I\|_F^2 + \|U^{(3)}U^{(3)T} - I\|_F^2), \tag{15}$$

$$L_{R_4}(U^{(4)}) = \frac{\rho}{R_4}(\|U^{(4)T}U^{(4)} - I\|_F^2 + \|U^{(4)}U^{(4)T} - I\|_F^2), \tag{16}$$

where ρ denotes regularization strength, ‖·‖$_F$ is the Frobenius norm. Orthogonality of the factor matrix $U^{(3)}$ is expressed by minimizing the residual matrix $L_{R_3}$. If $U^{(3)}$ exhibits strong orthogonality, then $L_{R_3}$ will be small. Therefore, the following loss can be modeled for the model.

$$L = L_{ce} + \lambda(L_{R_3}(U^{(3)}) + L_{R_4}(U^{(4)})), \tag{17}$$

where $L_{ce}$ is loss function, e.g. cross-entropy loss in classification tasks, and the second term is regularization loss. λ is a weight parameter. The whole training process is shown in Algorithm 1.

*E. Rank Selection*

In tensor decomposition, the choice of rank is crucial for model compression because it directly affects the accuracy and efficiency of the compressed model. The rank is the dimension of the core tensor in the decomposed tensor, and a lower rank means fewer parameters, which can reduce the storage requirements and computational complexity of the model. However, too low a rank may lead to information loss and affect the prediction accuracy of the model.

In this paper, VBMF (Variational Bayesian Matrix Factorization) is used to estimate the matrix or tensor rank. VBMF is based on Bayesian inference, which estimates the rank of a matrix or tensor by introducing a prior distribution and is particularly suitable for high-dimensional data. The basic idea of VBMF is to decompose the matrix or tensor into a low-rank part and a noise part and then use Bayesian inference to estimate the parameters of these parts, including the rank of the low-rank part. The VBMF method automatically adjusts the complexity of the model and takes into account the noise in the data. In VBMF, for a given matrix or tensor, variational inference can be used to estimate the parameters of its low-rank and noise parts. In this case, the estimation of the rank is done by maximizing the posterior probability. By iteratively updating the parameters, the rank value that maximizes the a posteriori probability can be found.

VBMF can efficiently estimate the tensor rank. However, for CNNs, there is a significant imbalance in the weight tensor, i.e., the input and output channels may not be equal. Decomposing different modalities using the same set of tensor ranks usually results in a severe loss of accuracy. Meanwhile, the potential relationship between the input and output channels, such as the multiplicative relationship, is taken into account. Therefore, the VBMF is only used to determine the rank of the third modal matrix. The rank of the fourth modal matrix can be obtained simply from the relationship between the channels.

TABLE I STATISTICAL INFORMATION ON CIFAR10 AND SVHN

| Datasets | class | Train | Test | Size |
|---|---|---|---|---|
| D1 | 10 | $5 \times 10^4$ | $1 \times 10^4$ | $32 \times 32$ |
| D2 | 100 | $5 \times 10^4$ | $1 \times 10^4$ | $32 \times 32$ |

IV. EXPERIMENTS

We tested the effectiveness of the method by testing ResNet18 and ResNet20 on the CIFAR-10 and CIFAR-100 datasets, respectively.

*A. Data preparation*

Different datasets will be used to evaluate the performance of the compression model. The informative statistics of the different datasets are displayed in Table I.

**D1(CIFAR-10)**：It is a widely used standard dataset for computer vision for training and evaluating machine learning models, especially in the field of image recognition.

**D2(CIFAR-100)**：It is a real image dataset widely used in the development of machine learning and image classification algorithms. The CIFAR-100 dataset is commonly used in deep learning research, especially in the field of CNNs and unsupervised feature learning.

*B. Evaluation Metrics*

**Top-1 accuracy (Top-1):** It is the proportion of the model's predictions for the most likely category (i.e., the prediction with the highest probability) that agrees with the true label. In other words, a prediction is considered correct if the model predicts exactly the right category as its first choice. Specific definitions are given below:

$$\text{Top-1} = \frac{f}{F} \times 100\%, \tag{18}$$

where $F$ denotes the total number of test samples. $f$ denotes the number of correctly predicted samples, i.e., the number of samples for which the model's Top-1 prediction matches the actual label.

**Compression ratio:** The compression ratio of a neural network is the ratio of the compressed model parameters to the original model parameters. It is defined as follows:

$$\text{CR} = \frac{P_{original}}{P_{compressed}}, \tag{19}$$

Where $P_{compressed}$ and $P_{original}$ denote the parameters after compression and before compression, respectively.

*C. Experimental platforms and setups.*

The experiments are conducted using PyTorch within the Tensorly toolbox. These experiments run on a desktop computer equipped with 2.50 GHz Intel Cores and 12 GB of RAM. The batch size is set to 128, and the initial learning rate is configured at 0.1. This rate is then decreased to 0.01 after the 100th epoch and further reduce to 0.001 after the 150th epoch.

*D. Experimental results analysis*

Through extensive experiments, we obtained the results shown in Tables II, III, IV, and V.

**ResNet20 on CIFAR-10.** Table II compares our method TK-2 format ResNet-20 models with the state-of-the-art model compression method on the CIFAR-10 dataset. Some sparse compression methods, such as GrowEfficient and BackSparse, introduce a serious performance degradation to the model, even though the compression is relatively small. SVDT performs similarly on the CIFAR-10 dataset. In contrast, the PSTRN-S method obtains 0.19% accuracy improvement with a similar compression ratio 2.70×. It is exciting that our approach achieves state-of-the-art results, both in terms of accuracy and compression rate. Especially when the compression ratio is small, our method can obtain an accuracy improvement of 0.22% with a moderate compression ratio 2.00× compared to the original model.

**ResNet20 on CIFAR-100.** Table III shows the experimental results on CIFAR-100. Our compression model outperforms other work when the compression ratio is small. With 2.30× compression ratio, our model achieves 66.96% Top-1 accuracy, which is even 1.56% higher than the uncompressed model. If the compression ratio exceeds 4, the accuracy of the model suffers a severe degradation, both for the state-of-the-art BATUDE and the method proposed in this paper. Even so, the method in this paper maintains sufficient toughness in terms of comprehensive performance.

**ResNet32 on CIFAR-10.** Table VI shows the experimental results on CIFAR-10. ResNet32 has a deeper residual structure compared to ResNet20, so the classification accuracy obtained will be higher. Similar to the previous results, utilizing sparsity for model compression brings a non-negligible accuracy loss, even if the compression is relatively small. For some advanced low-rank models such as PSTRN-S, it has 1.05% accuracy drop with compression ratio 2.60×. However, our model even outperforms the uncompressed model with 0.5% accuracy increase.

**ResNet32 on CIFAR-100.** Table V shows the experimental results on CIFAR-100. Again, our approach still achieves state-of-the-art performance. The loss of accuracy is more pronounced in both PSTRN-M and PSTRN-S. In contrast, BATUDE, a low-rank decomposition model, exhibits strong compression. Similar results are achieved with our method when the compression is relatively large. However, when the compression ratio is small, our method is more advantageous. Specifically, our method can obtain an accuracy improvement of 0.26% with a moderate compression ratio 2.00× compared to the original model.

TABLE II RESNET20 ON CIFAR-10 DATASET

| Method | Post-Train Model | Top-1 (%) | CR |
|---|---|---|---|
| Original ResNet-20 | Dense | 91.25 | 1.00× |
| SVDT [91] | Low-rank | 90.39 | 2.94× |
| PSTRN-S [92] | Low-rank | 91.44 | 2.70× |
| GE [93] | Sparse | 90.91 | 2.00× |
| BS [94] | Sparse | 90.73 | 2.77× |
| **Ours** | Low-rank | 91.79 | 2.00× |
| **Ours** | Low-rank | 89.42 | 4.26× |

TABLE III RESNET20 ON CIFAR-100 DATASET

| Method | Post-Train Model | Top-1 (%) | CR |
|---|---|---|---|
| Original ResNet-20 | Dense | 65.40 | 1.00× |
| PSTRN-M [92] | Low-rank | 63.62 | 4.70× |
| PSTRN-S [92] | Low-rank | 66.13 | 2.30× |
| BATUDE [34] | Low-rank | 64.91 | 4.70× |
| BATUDE [34] | Low-rank | 66.67 | 2.80× |
| **Ours** | Low-rank | 66.96 | 2.00× |
| **Ours** | Low-rank | 65.12 | 4.26× |

TABLE VI RESNET32 ON CIFAR-10 DATASET

| Method | Post-Train Model | Top-1 (%) | CR |
|---|---|---|---|
| Original ResNet-32 | Dense | 92.49 | 1.00× |
| FPGM [95] | Sparse | 91.93 | 2.12× |
| SCOP [96] | Sparse | 92.13 | 2.27× |
| SVDT [91] | Low-rank | 90.55 | 3.93× |
| PSTRN-S [92] | Low-rank | 91.44 | 2.60× |
| **Ours** | Low-rank | 92.99 | 2.00× |
| **Ours** | Low-rank | 91.07 | 4.26× |

TABLE V RESNET32 ON CIFAR-100 DATASET

| Method | Post-Train Model | Top-1 (%) | CR |
|---|---|---|---|
| Original ResNet-32 | Dense | 68.10 | 1.00× |
| PSTRN-M [92] | Low-rank | 59.03 | 2.50× |
| PSTRN-S [92] | Low-rank | 66.77 | 5.20× |
| BATUDE [34] | Low-rank | 68.05 | 2.40× |
| BATUDE [34] | Low-rank | 66.96 | 5.20× |
| **Ours** | Low-rank | 68.37 | 2.00× |
| **Ours** | Low-rank | 67.01 | 4.26× |

## V. CONCLUSIONS

In this paper, we propose a model compression framework Based on low-rank representation learning that incorporates orthogonal regularization and VBMF. In this framework, over-parameterization and orthogonal regularization training provide orthogonal properties for low-rank models. In addition, the VBMF estimates the rank of one modality of the weight tensor, and the other rank is solved taking into account the imbalance between channels. It avoids that one modality of the tensor affects the model performance because it is over-compressed. Experiments show that the method in this paper outperforms the state-of-the-art models in terms of compression rate and test accuracy.

In the future, we will design novel tensor decomposition methods to achieve a balance between model parameters and accuracy. Meanwhile, to increase the compression rate of the model, we will synergistically explore different compression methods, such as low-rank decomposition combined with pruning and quantization.